\documentclass[10pt,twocolumn,letterpaper]{article}
\usepackage{con}
\usepackage{times}
\usepackage{epsfig}
\usepackage{graphicx}
\usepackage{amsmath}
\usepackage{amssymb}
\usepackage{colortbl}
\usepackage{booktabs}  
\usepackage{threeparttable} 
\usepackage{multirow} 
\usepackage[table,xcdraw]{xcolor}
\usepackage{pifont}
\usepackage{bbding}
\usepackage[pagebackref=true,breaklinks=true,colorlinks,bookmarks=false]{hyperref}

\usepackage[breaklinks=true,bookmarks=false]{hyperref}

\iccvfinalcopy 

\ificcvfinal\fi

\begin{document}

\title{TokenPose: Learning Keypoint Tokens for Human Pose Estimation}

\author{Yanjie Li\thanks{This work was done when Yanjie and Sen Yang were interns at MEGVII Tech.}$^{* 1,2}$\quad Shoukui Zhang$^{2}$\quad Zhicheng Wang$^{2}$\quad Sen Yang$^{* 2,3}$\\ Wankou Yang$^{3}$\quad Shu-Tao Xia\thanks{Corresponding author.}\textsuperscript{$\dagger 1,4$}\quad Erjin Zhou$^{2}$\\[1.mm]
$^{1}$Tsinghua Shenzhen International Graduate School, Tsinghua University\\ $^{2}$MEGVII Technology \quad $^{3}$Southeast University\\
$^{4}$PCL Research Center of Networks and Communications, Peng Cheng Laboratory\\
{\tt\small lyj20@mails.tsinghua.edu.cn}\quad{\tt\small \{zhangshoukui, wangzhicheng\}@megvii.com}\\ {\tt\small \{yangsenius, wkyang\}@seu.edu.cn}\quad{\tt\small xiast@sz.tsinghua.edu.cn}\quad{\tt\small zej@megvii.com}
}

\maketitle

\ificcvfinal\thispagestyle{empty}\fi
\begin{abstract}
 Human pose estimation deeply relies on visual clues and anatomical constraints between parts to locate keypoints. Most existing CNN-based methods do well in visual representation, however, lacking in the ability to explicitly learn the constraint relationships between keypoints. In this paper, we propose a novel approach based on Token representation for human Pose estimation~(TokenPose). In detail, each keypoint is explicitly embedded as a token to simultaneously learn constraint relationships and appearance cues from images. Extensive experiments show that the small and large TokenPose models are on par with state-of-the-art CNN-based counterparts while being more lightweight. Specifically, our TokenPose-S and TokenPose-L achieve $72.5$ AP and $75.8$ AP on COCO validation dataset respectively, with significant reduction in parameters (\textcolor{red}{ $\downarrow80.6\%$} ; \textcolor{red}{$\downarrow$ $56.8\%$}) and GFLOPs (\textcolor{red}{ $\downarrow$ $75.3\%$}; \textcolor{red}{$\downarrow$ $24.7\%$}). Code is publicly available\footnote{\url{https://github.com/leeyegy/TokenPose}}.
\end{abstract}

\section{Introduction}

$2$D human pose estimation aims to localize human anatomical keypoints which deeply relies on both visual cue and keypoints constraint relationships. It is a fundamental task in computer vision, which has attracted extensive attention from academia and industry.

Over the past decade, deep convolutional neural networks have achieved impressive performances on human pose estimation due to their powerful capacity in visual representation and recognition~\cite{20201207HumanPose-HigherHRNet,20201207-human_pose-hrnet,20201207-HumanPose-hourglass,20201209-Human_pose-asso_embedding,ECCV18-xiao-simple,rmpe,wei2016convolutional,rmi}. Since \emph{heatmap representation} has become the standard label representation to encode the positions of keypoints, most existing models tend to use fully convolutional layers to maintain the $2$D-structure of feature maps until the network output. Nevertheless, there are usually no concrete variables abstracted by such CNN models to directly represent the keypoint entities, which limits the ability of the model to explicitly capture constraint relationships between parts.

\begin{figure}[t]
    \centering
    \vspace{-2em}
    \includegraphics[width=0.5\textwidth]{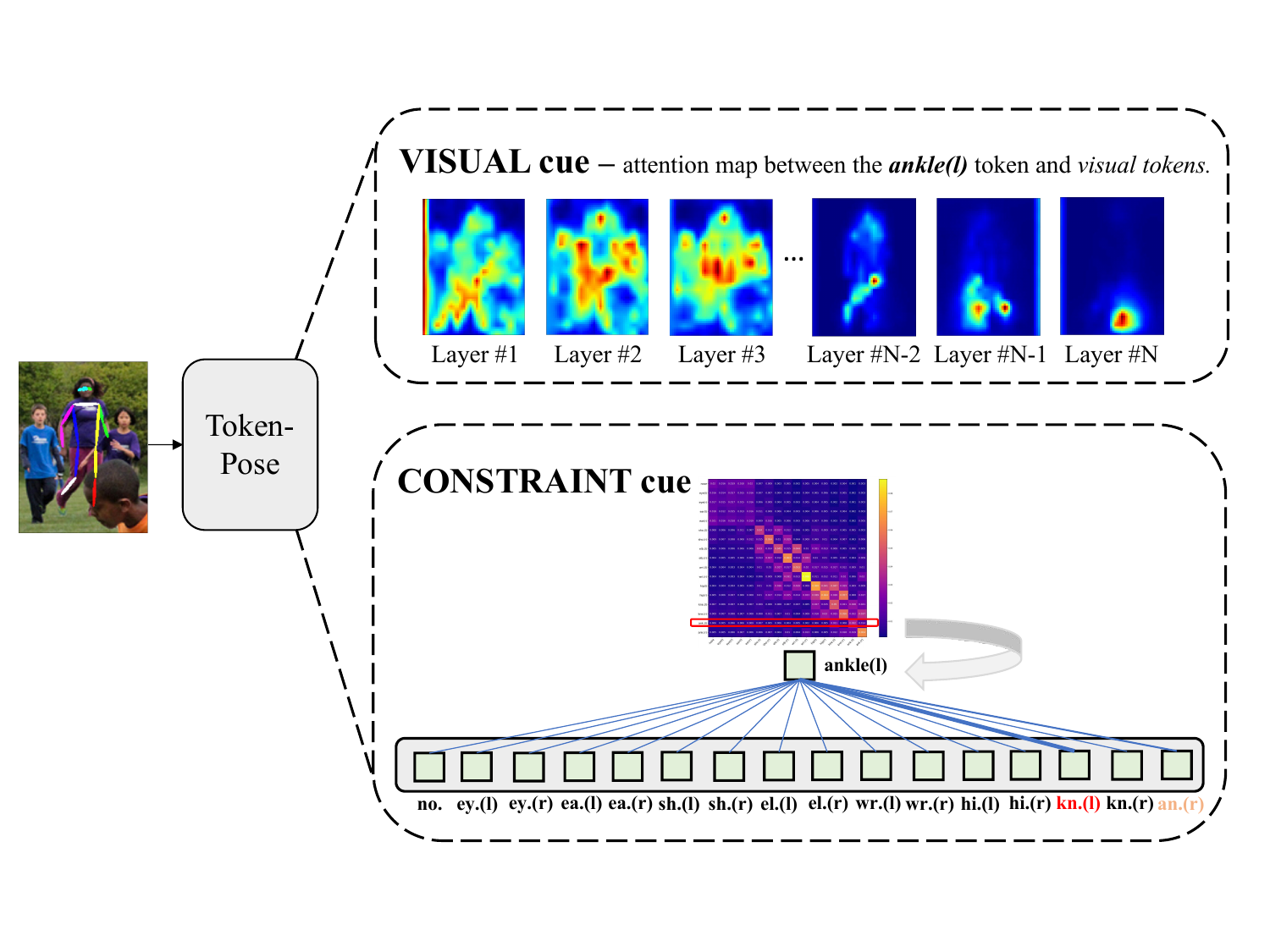}
    \vspace{-1.0cm}
    \caption{The process of predicting the location of the \textbf{\emph{left ankle}}. For visual cue learning, the proposed TokenPose focuses on the global context in the first few layers, and then gradually converges to some local regions as the network goes deeper. In the last few layers, TokenPose has considered \emph{hip} and \emph{knee} in turn which are close to the target keypoint, and finally localizes the position of the \textbf{\emph{left ankle}}. For constraint cue learning, TokenPose shows that localizing the \textbf{\emph{left ankle}} mostly relies on the \emph{left knee} and \emph{right ankle}, corresponding to \textbf{\emph{adjacency}} constraint and \textbf{\emph{symmetric}} constraint respectively.}
    \label{fig:motivation}
    \vspace{-0.4em}
\end{figure}

Recently, Transformer~\cite{transformer} and its variants that originated from natural language processing (NLP) have merged as new choices for various vision tasks. Its ability of modeling global dependencies is more powerful than CNN, which points out a promising way to efficiently capture relationships between visual entities/elements. And in the field of NLP, all language elements such as words or characters are usually symbolized by embeddings or token vectors with fixed dimensions, so as to better measure their similarities in a vector space, like the way of word2vec~\cite{mikolov2013efficient}. 

\begin{figure*}[ht]
    \centering
    \vspace{-2em}
    \includegraphics[width=0.95\textwidth]{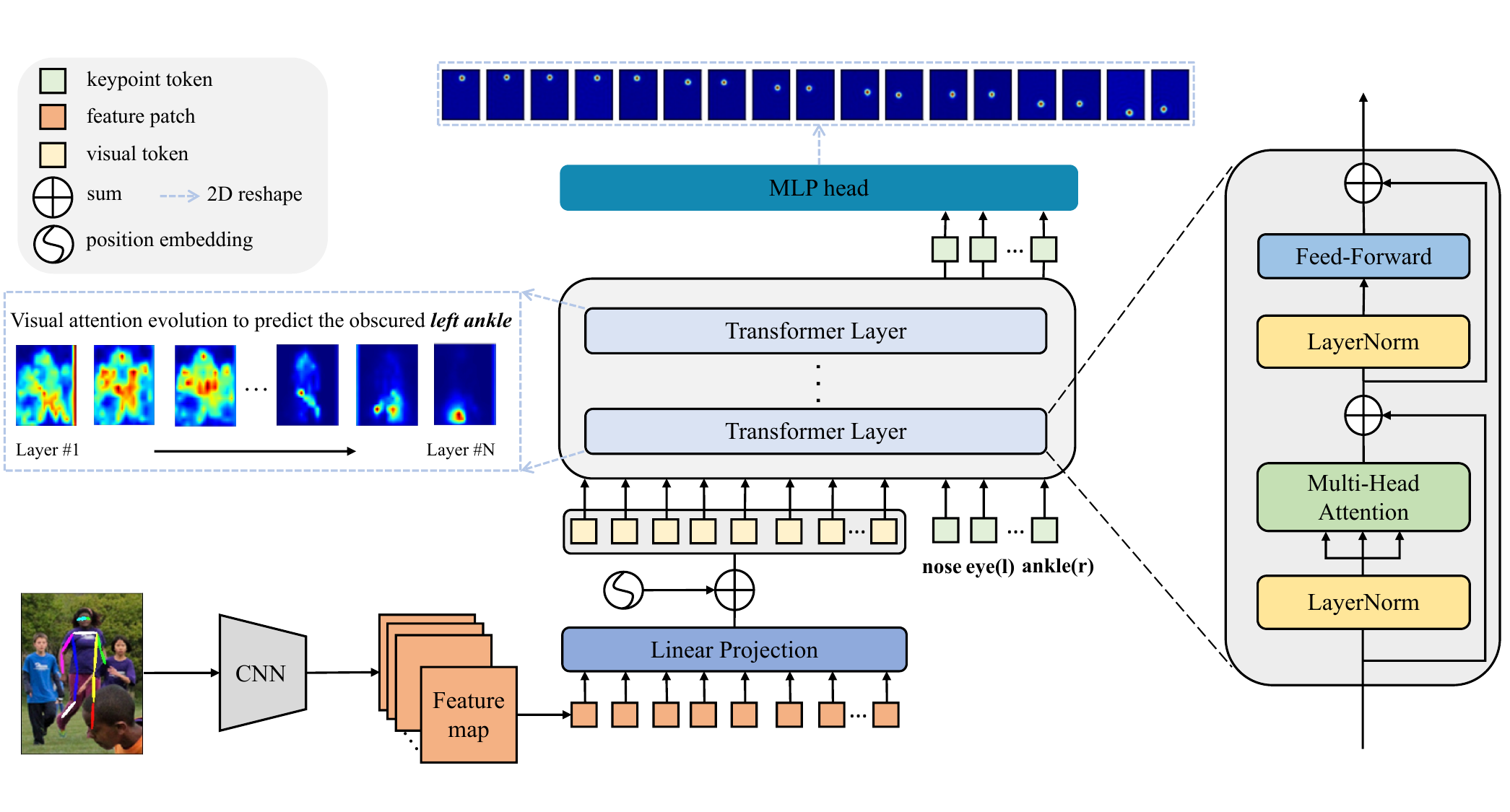}
    \vspace{-0.3cm}
    \caption{Schematic illustration of the proposed TokenPose. The feature maps extracted by CNN backbone are uniformly split into patches and flattened to $1$D vectors. Visual tokens are yielded by adopting a linear projection to embed the flattened vectors. In addition, keypoint tokens are initialized randomly to represent each specific type of keypoint. Then, the $1$D sequence of visual tokens and keypoint tokens are taken as input to Transformer encoder. Both appearance cues and anatomical constraint cues are captured through self-attention interactions in each Transformer layer. Finally, the keypoint tokens outputted by the last Transformer layer are used to predict the keypoints heatmaps via an MLP head.}
    \label{fig:model}
    \vspace{-0.1em}
\end{figure*}

We borrow such a concept of  ``token'' and present a novel token-based representation for human pose estimation, namely TokenPose. Specifically, we conduct two different types of tokenizations: keypoint tokens and visual tokens. Visual tokens are yielded by uniformly splitting an image into patches and mapping the flattened patches into embeddings with fixed dimensions. Meanwhile, keypoint tokens are randomly initialized embeddings, each of which represents a specific type of keypoint (e.g., left knee, left ankle, right eye, etc.). The resulting keypoint tokens can learn both visual clues and constraint relations from interactions with visual tokens and the other keypoint tokens respectively. An example of how the proposed model predicts the location of \emph{left ankle} is shown in Figure~\ref{fig:motivation}. The positions of keypoints are finally estimated over the token-based representation outputted by our network. The architecture of TokenPose is illustrated in Figure~\ref{fig:model}. 

It is worth noting that TokenPose learns the statistic constraint relationships between keypoints from large amounts of data. Such information is encoded into keypoint tokens that can record their relationships by vector similarities. During inference, TokenPose associates keypoint tokens with those visual tokens whose corresponding patches possibly contain the target keypoints. By visualizing the attentions, we can observe how they interact and how the model exploits cues to localize keypoints.

The contributions are summarized as follows:
\begin{itemize}
    \item We propose to use \emph{token} to represent each \emph{keypoint entity}. In this way, visual cue and constraint cue learning are explicitly incorporated into a unified framework. 
    \vspace{-0.5em}
    \item Both hybrid and pure Transformer-based architectures are explored in this work. As far as we know, proposed TokenPose-T is the first pure Transformer-based model for $2$D human pose estimation.
    \vspace{-0.5em}
    \item We conduct experiments over two widely-used benchmark datasets: COCO keypoint detection dataset~\cite{coco} and MPII Human Pose dataset~\cite{mpii}. TokenPose achieves competitive state-of-the-art performance with much fewer parameters and computation cost compared with existing CNN-based counterparts.
\end{itemize}

\section{Related Work}
\subsection{Human Pose Estimation}
 Deep convolutional neural networks have been applied to human pose estimation which greatly boost the model performance~\cite{tang2018deeply,gkioxari2016chained,20201207-human_pose-hrnet,ECCV18-xiao-simple,20201207-HumanPose-hourglass,voting,20201209-Human_pose-asso_embedding,cai2020learning, cvpr18-cascaded}.
 
 Recent heatmap-based methods tend to improve performance by stacking deeper network architecture. Hourglass~\cite{20201207-HumanPose-hourglass} stacks blocks to enhance the heatmap estimation quality. SimpleBaseline~\cite{ECCV18-xiao-simple} designs a simple architecture by stacking transposed convolution layers and achieves impressive performances. HRNet~\cite{20201207-human_pose-hrnet} proposes to maintain high-resolution representation through the whole process in order to provide spatially precise heatmap estimation. However, it is still hard for convolutional neural networks to capture and model constraint relationships between keypoints, which are important for human pose estimation.
 
\subsection{Vision Transformer}
Transformer~\cite{transformer} adopts encoder-decoder architecture based on self-attention and feed-forward network, which achieves great success in NLP. Recently, Transformer-based models~\cite{vit,deit,detr,tnt,setr,zhu2020deformabledetr,dai2020up,yang2020transpose,chen2020pretrained,pvt,yuan2021tokens, stoffl2021end} have also shown enormous potential in various vision tasks. 
\paragraph{Detection.}DETR~\cite{detr} proposes a Transformer based architecture to handle object detection end-to-end, effectively eliminating the need for many hand-designed components. Deformable DETR~\cite{zhu2020deformabledetr} then proposes to make attention modules only attend to a small set of key sampling points around a reference, achieving better performance than DETR. UP-DETR~\cite{dai2020up} unsupervisedly pre-train DETR by design randomly cropped patches.

\paragraph{Classification.}
ViT~\cite{vit} proposes a pure Transformer model with patch embedding representation, which is pre-trained on large amounts of data and then fine-tuned on ImageNet dataset. DeiT~\cite{deit} introduces a distillation token to ViT to learn knowledge from a teacher network, to avoid the pre-training on a large dataset. Tokens2Token~\cite{yuan2021tokens} progressively encodes image into tokens and models the local structure information to reduce the sequence length.

\paragraph{Human Pose Estimation.} Recent several works~\cite{15keypoints,he2020epipolar, lin2020end,yang2020transpose,zheng20213d,stoffl2021end} introduce Transformer for human pose estimation. PoseFormer~\cite{zheng20213d} introduces Transformer for $3$D pose estimation, based on $2$D pose sequences in video frames. TransPose~\cite{yang2020transpose} tends to utilize attention layers built in Transformer to reveal the long-range dependencies of the predicted keypoints. However, TransPose lacks the ability to directly model the constraint relationships between keypoints. In this work, we propose to explicitly represent keypoints as token embeddings. And then both visual clues and constraint relations are simultaneously learned through self-attention interactions.

\section{Method}
We firstly revisit the heatmap-based Fully Convolutional Networks (FCNs) for human pose estimation, and then describe our token-based design.
\subsection{FCN-based Human pose estimation}
The goal of human pose estimation is to localize $N$ keypoints or parts from an image $I$ with size $H\times W\times 3$. Nowadays, heatmap-based fully convolutional neural networks~\cite{wei2016convolutional,cai2020learning,20201207-HumanPose-hourglass,yang2017learning,cvpr18-cascaded,papandreou2018personlab,ECCV18-xiao-simple,20201207-human_pose-hrnet} have been dominant solutions due to their high performance.

The widely-adopted pipeline is to utilize convolutional neural network to yield multi-resolution image feature maps, and a regressor to estimate $N$ heatmaps of size $\hat{H}\times\hat{W}$. In order to yield $N$ heatmaps, a 1$\times$1  convolutional layer tends to be adopted to quickly adapt the channels of feature maps to $N$.

Despite the great success existing FCN-based methods have achieved, it's tough for CNN to explicitly capture constraint relationships between keypoints, which results in sub-optimal model design for this task. 

\begin{table*}\large

\begin{center}
\scalebox{0.75}{
\setlength{\tabcolsep}{1mm}
\begin{tabular}{l|c|cccc|cc}
\toprule
Model & CNN backbone & Layers & Embedding size & Heads & Patch size & \#Params & GFLOPs \\ \midrule
TokenPose-Tiny & - & $12$ & $192$ & $16$ & $16\times12$ & $5.8$M & $1.3$ \\
TokenPose-Small-v$1$ & stem-net & $12$ & $192$ & $8$ & $4\times3$ & $6.6$M & $2.2$ \\
TokenPose-Small-v$2$ & stem-net & $12$ & $192$ & $8$ &  $2\times2$ & $6.2$M & $11.6$ \\
TokenPose-Base & HRNet-W$32$-stage$3$ & $12$ & $192$ & $8$ & $4\times3$ &$13.5$M  & $5.7$  \\
TokenPose-Large/D$6$ & HRNet-W$48$-stage$3$ & $6$ & $192$ & $8$ & $4\times3$ & $20.8$M & $9.1$ \\ 
TokenPose-Large/D$24$ & HRNet-W$48$-stage$3$ & $24$ & $192$ & $12$ & $4\times3$ & $27.5$M & $11.0$ \\
\bottomrule
\end{tabular}
}
\end{center}
\caption{Architecture configurations. The model parameters and GFLOPs are computed under an image with $256\times192$ input resolution.}
\label{tab:variant}
\end{table*}

\subsection{Token-based Keypoint Representation}
\paragraph{Visual tokens.}
The standard Transformer~\cite{transformer} accepts a $1$D sequence of token embeddings as input. To handle $2$D images, we follow the process of ViT~\cite{vit}. An image $x \in \mathbb{R}^{H \times W \times C}$ 
is divided into a grid of $\frac{H}{P_h}\times \frac{W}{P_w}$ patches uniformly of size $P_h \times P_w$. And then each patch $p$ is flattened into a 1D vector with size of $P_h\cdot P_w \cdot C$. To obtain a visual token $v$, each flattened patch $p$ is then mapped into a $d$-dimensional embedding by a linear projection function $f: p \longrightarrow v\in \mathbb{R}^d$. 

Considering human pose estimation is a location-sensitive vision task, $2$D position embedding~\cite{transformer} $pe_i$ is added to every specific visual token $v_i$ to produce the input visual tokens \verb|[visual]|$=\{v_1+pe_1, v_2+pe_2, ... , v_L+pe_L\}$, where $L=\frac{H\times W}{P_h\times P_w}$ is the amount of visual tokens .  In this way, each visual token is yielded to represent a specific area of original image.
\paragraph{Keypoint tokens.} We prepend $N$ learnable $d$-dimensional embedding vectors to represent $N$ target keypoints. We symbolize the keypoint tokens as \verb|[keypoint]|. Together with visual tokens processed from image patches, keypoint tokens are accepted as the input of the Transformer. The state of $N$ keypoint tokens at the output of the Transformer encoder serves as the $N$ keypoints representation. 

\paragraph{Transformer.} Given the $1$D token embeddings sequence $T=\{\verb|[visual]|,\verb|[keypoint]|\}$ as input, the Transformer encoder~\cite{transformer} learns keypoint feature representation by stacking $M$ blocks. Each block contains a Multi-head Self-attention (MSA) module and a Multilayer Perceptron (MLP) module. In addition, layernorm (LN) is adopted before every module. Self-attention (SA) can be formulated as:
\begin{equation}
    SA(T^{l-1}) = softmax(\frac{T^{l-1}W_Q(T^{l-1}W_K)^T}{\sqrt{d_h}})(T^{l-1}W_V)
\end{equation}
where $W_Q, W_K, W_V\in\mathbb{R}^{d\times d}$ are the learnable parameters of three linear projection layers, $T^{l-1}$ is the output of the $(l-1)$-th layer, $d$ is the dimension of tokens, and $d_h=d$. MSA is an extension of SA with $h$ self-attention operations which are called ``heads". In MSA, $d_h$ is typically set to $d/h$.
\begin{equation}
    MSA(\textbf{T}) = [SA_1(T); SA_2(T); ... ; SA_h(T)]W_P
\end{equation}
where $W_P\in \mathbb{R}^{(h\cdot d_h)\times d}$. Note, the final heatmap prediction is based on the \verb|[keypoint]| tokens outputted by the Transformer encoder with $M$ blocks, which are denoted as $\{T_1^M, T_2^M, ... , T_N^M\}$.

\paragraph{Heatmap estimation.} To obtain the $2$D heatmaps with size of $\hat{H}\times \hat{W}$, the $d$-dimensional \verb|[keypoint]| tokens outputted by the Transformer encoder are mapped into $\hat{H}\cdot\hat{W}$-dimensional feature vectors by linear projection. The mapped $1$D vectors are then reshaped to $2$D heatmaps. In addition, the MSE loss function is adopted to compare the predicted heatmaps and the groundtruth heatmaps.

\paragraph{Hybrid architecture.} Instead of manipulating raw image patches directly, the input visual tokens can also be extracted from feature maps outputted by a convolution neural network~\cite{CNN}. In the hybrid architecture, CNN is adopted to extract low-level image features more efficiently. 

\section{Experiments}
\subsection{Model Variants}
We provide both hybrid and pure Transformer-based variants for TokenPose. For hybrid architecture, convolutional neural networks with various depths are used for image feature extracting. The configuration details are presented in Table~\ref{tab:variant}. Note, TokenPose-T* is the pure Transformer-based variant. TokenPose-S*, TokenPose-B and TokenPose-L* adopt stem-net\footnote{It's widely used to quickly downsample the feature map into $1/4$ input resolution, consisting of a very shallow convolutional structure\cite{20201207-human_pose-hrnet,20201207HumanPose-HigherHRNet}.}, HRNet-W32~\cite{20201207-human_pose-hrnet} and HRNet-W48~\cite{20201207-human_pose-hrnet}  as backbone, respectively.

In this paper, brief notation is used for convenience. For example, TokenPose-L/D$24$ means the ``Large" variant with $24$ Transformer layers. Unless noted otherwise, TokenPose-S and TokenPose-L are used as the abbreviations for TokenPose-Small-v$2$ and TokenPose-Large/D$24$.

\subsection{COCO Keypoint Detection}
\paragraph{Dataset.} The COCO dataset~\cite{coco} consists of more than $200,000$ images and $250,000$ person instances which are labeled with $17$ keypoints. The COCO dataset is divided into \emph{train/val/test-dev} sets, which contains $57$k, $5$k and $20$k images respectively. All the experiments reported in this paper are trained only on the train$2017$ set. The methods are evaluated on the val$2017$ set and test-dev$2017$ set.

\begin{table*}[]
\begin{center}
\scalebox{0.73}{
\renewcommand{\arraystretch}{1.1}
\setlength{\tabcolsep}{1.5mm}
\begin{tabular}{l|cc|c|l|l|c|cccccc}
\toprule
\multirow{2}{*}{Method} & \multicolumn{2}{c|}{Pretrain} & \multirow{2}{*}{Input size} & \multirow{2}{*}{\#Params} & \multirow{2}{*}{GFLOPs} & \multirow{2}{*}{gtbbox AP} & \multirow{2}{*}{$AP$} & \multirow{2}{*}{$AP^{50}$} & \multirow{2}{*}{$AP^{75}$} & \multirow{2}{*}{$AP^M$} & \multirow{2}{*}{$AP^L$} & \multirow{2}{*}{$AR$} \\ \cline{2-3}
 & CNN & Transformer &  &  &  &  &  &  &  &  &  &  \\ \hline
SimpleBaseline-Res$50$~\cite{ECCV18-xiao-simple} & Y & - & $256\times192$ &$34.0$M\dag & $8.9$\dag & $72.4$ & $70.4$ & $88.6$ & $78.3$ & $67.1$ & $77.2$ & $76.3$ \\\hline
SimpleBaseline-Res$101$~\cite{ECCV18-xiao-simple} & Y & - & $256\times192$ & $53.0$M & $12.4$& - & $71.4$ & $89.3$ & $79.3$ & $68.1$ & $78.1$ & $77.1$ \\
SimpleBaseline-Res$152$~\cite{ECCV18-xiao-simple} & Y & - & $256\times192$ & $68.6$M\ddag & $15.7$\ddag & $74.3$ & $72.0$ & $89.3$ & $79.8$ & $68.7$ & $78.9$ & $77.8$ \\\hline
HRNet-W$32$\cite{20201207-human_pose-hrnet} & Y & - & $256\times192$ & $28.5$M\S & $7.1$\S& $76.5$ & $74.4$ & $90.5$ & $81.9$ & $70.8$ & $81.0$ & $79.8$ \\
HRNet-W$48$~\cite{20201207-human_pose-hrnet} & Y & - & $256\times192$ & $63.6$M$\pitchfork$& $14.6$$\pitchfork$ & $77.1$ & $75.1$ & $\textbf{90.6}$ & $82.2$ & $71.5$ & $81.8$ & $80.4$ \\
\midrule
\textbf{TokenPose-T  (pure Transformer)} & - & N & $256\times192$ & $5.8$M & $1.3$ & - &$65.6$ & $86.4$ & $73.0$ & $63.1$ & $71.5$ & $72.1$ \\ \hline
\textbf{TokenPose-S-v}$\textbf{1}$ & N & N & $256\times192$ & $6.6$M$\dag$ (\textcolor{red}{$\downarrow80.6\%$}) & $2.2$$\dag$ (\textcolor{red}{$\downarrow75.3\%$}) & $75.0$& $72.5$ & $89.3$ & $79.7$ & $68.8$ & $79.6$ & $78.0$ \\ 
\textbf{TokenPose-S-v}$\textbf{2}$ & N & N & $256\times192$ & $6.2$M$\ddag$ (\textcolor{red}{$\downarrow91.0\%$}) & $11.6$$\ddag$ (\textcolor{red}{$\downarrow23.7\%$}) & $76.1$ & $73.5$ & $89.4$ & $80.3$ & $69.8$ & $80.5$ & $78.7$ \\ \hline
\textbf{TokenPose-B} & Y & N & $256\times192$ & $13.5$M$\S$ (\textcolor{red}{$\downarrow52.6\%$}) & $5.7$$\S$ (\textcolor{red}{$\downarrow19.7\%$})  & - & $74.7$ & $89.8$ & $81.4$ & $71.3$ & $81.4$ & $80.0$ \\
\textbf{TokenPose-L/D}$\textbf{6}$ & Y & N & $256\times192$ & $20.8$M$\pitchfork$ (\textcolor{red}{$\downarrow67.3\%$}) & $9.1$$\pitchfork$ (\textcolor{red}{$\downarrow37.7\%$}) & $77.7$ & $75.4$ & $90.0$ & $81.8$ & $71.8$ & $82.4$ & $80.4$ \\
\textbf{TokenPose-L/D}$\textbf{24}$& Y & N & $256\times192$ & $27.5$M$\pitchfork$ (\textcolor{red}{$\downarrow56.8\%$}) & $11.0$$\pitchfork$ (\textcolor{red}{$\downarrow24.7\%$}) & $\textbf{78.2}$ & $\textbf{75.8}$ & $90.3$ & $\textbf{82.5}$ & $\textbf{72.3}$ & $\textbf{82.7}$ & $\textbf{80.9}$ \\
\bottomrule
\end{tabular}
}
\end{center}
\caption{Comparisons on the COCO validation set, provided with the same detected human boxes. Pretrain means pre-training the corresponding parts on the ImageNet classification task. TokenPose-S*, TokenPose-B* and TokenPose-L* achieve competitive results to SimpleBaseline~\cite{ECCV18-xiao-simple} and HRNet~\cite{20201207-human_pose-hrnet} respectively, with much fewer parameters\&GFLOPs. We compute the percentages in terms of parameters\&GFLOPs reduction between models marked with the same symbol.}
\label{tab:val}
\end{table*}

\paragraph{Evaluation metric.} We adopt standard average precision (AP) as our evaluation metric on the COCO dataset. AP is calculated based on Object Keypoint Similarity (OKS): $OKS=\frac{\sum_i exp(-\hat{d}_i^2/2s^2k_i^2)\sigma(v_i>0)}{\sum_i \sigma(v_i>0)}$, where $\hat{d}_i$ is the Euclidean distance between the $i$-th predicted keypoint coordinate and the corresponding groundtruth, $v_i$ is the visibility flag of the keypoint, $s$ is the object scale, and $k_i$ is a keypoint-specific constant.   

\paragraph{Baseline settings.} For model training, we use the Adam optimizer. For HRNet~\cite{20201207-human_pose-hrnet} and SimpleBaseline~\cite{ECCV18-xiao-simple}, we simply follow the original settings in their paper. 

\paragraph{Implementation details.} In this paper, we follow the two-stage top-down human pose estimation paradigm similar to \cite{20201207-human_pose-hrnet,cvpr18-cascaded,ECCV18-xiao-simple,mppe}. In the paradigm,  the single person instance is firstly detected by a person detector, and then keypoints are predicted. We adopt the widely-used person detectors provided by SimpleBaseline~\cite{ECCV18-xiao-simple} on the validation set and test-dev set. To alleviate the quantisation error, the well-designed coordinate decoding strategy~\cite{darkpose} is adopted.

For our work, the base learning rate is set as $1$e-$3$, and is dropped to $1$e-$4$ and $1$e-$5$ at the $200$th and $260$th epochs, respectively. The total training process requires $300$ epochs, given Transformer structure tends to rely on longer training to convergence. We follow the data augmentation in~\cite{20201207-human_pose-hrnet}. 

\begin{table}[]
\begin{center}
\resizebox{1\columnwidth}{!}{
\begin{tabular}{l|cc|c|c}
\toprule
Model & \multicolumn{1}{l}{Embedding size} & \multicolumn{1}{l|}{Layer} & AP & \#Params \\ \hline
TokenPose-L/D$12$ & $192$ & $12$ & $75.3$  & $23.0$M \\
TokenPose-L/D$24$ & $192$ & $24$ & $75.8$ & $27.5$M \\
TokenPose-L+/D$12$ & $384$ & $12$ & $75.5$ & $38.2$M \\
\bottomrule
\end{tabular}
}
\end{center}
\caption{Results of model scaling on the COCO validation set. The input image size is $256\times192$.}
\label{tab:scaling}
\end{table}
\paragraph{Comparison with state-of-the-art methods.} As Table~\ref{tab:val} shown, our proposed TokenPose achieves competitive performance compared with the other state-of-the-art methods via much fewer model parameters and GFLOPs. Compared to SimpleBaseline~\cite{ECCV18-xiao-simple} that adopts ResNet-50 as the backbone, our TokenPose-S-v1 improves AP by $2.1$ points with significant reduction in both model parameters ($\downarrow 80.6\%$) and GFLOPs ($\downarrow 75.3\%$). Compared to SimpleBaseline~\cite{ECCV18-xiao-simple} that uses ResNet-$152$ as the backbone, our TokenPose-S-v2 achieves better performance, while using only $9.0\%$ model parameters. Compared with HRNet-W$32$ and HRNet-W$48$, TokenPose-B and TokenPose-L achieve similar results with less than $50\%$ model parameters, respectively. Besides, TokenPose-T obtains $65.6$ AP with only $5.8$M model parameters and $1.3$ GFLOPs, without any convolution layer. Note, all Transformer parts are trained from scratch, without any pre-training. Also, Table~\ref{tab:test-dev} shows the results of our method and the existing state-of-the-art methods on the COCO test-dev set. With $384\times288$ as the input resolution, our TokenPose-L/D$24$ achieves $75.9$ AP.

\begin{table}[t]
\begin{center}
\resizebox{\columnwidth}{!}{
\renewcommand{\arraystretch}{1.1}
\setlength{\tabcolsep}{1.5mm}
 
\begin{tabular}{l|cccccccc|c}
\toprule
Model & Hea & Sho & Elb & Wri & Hip & Kne & Ank & Mean & \#Params  \\ \hline
SimpleBaseline-Res$50$~\cite{ECCV18-xiao-simple} & $96.4$ & $95.3$ & $89.0$ & $83.2$ & $88.4$ & $84.0$ & $79.6$ & $88.5$ & $34.0$M \\
SimpleBaseline-Res$101$~\cite{ECCV18-xiao-simple} & $96.9$ & $95.9$ & $89.5$ & $84.4$ & $88.4$ & $84.5$ & $80.7$ & $89.1$ & $53.0$M \\
SimpleBaseline-Res$152$~\cite{ECCV18-xiao-simple} & $97.0$ & $95.9$ & $90.0$ & $85.0$ & $89.2$ & $85.3$ & $81.3$ & $89.6$ & $68.6$M \\
HRNet-W$32$~\cite{20201207-human_pose-hrnet} & $96.9$ & $\textbf{96.0}$ & $90.6$ & $85.8$ & $88.7$ & $86.6$ & $82.6$ & $90.1$ & $28.5$M \\\hline
TokenPose-L/D$6$ & $97.1$ & $95.9$ & $\textbf{91.0}$ & $85.8$ & $\textbf{89.5}$ & $86.1$ & $\textbf{82.7}$ & $\textbf{90.2}$ & $\textbf{21.4M}$\\
TokenPose-L/D$12$ & $\textbf{97.2}$ & $95.8$ & $90.7$ & $85.9$ & $89.2$ & $86.2$ & $82.3$ & $90.1$ & $23.5$M  \\
TokenPose-L/D$24$ & $97.1$ & $95.9$ & $90.4$ & $\textbf{86.0}$ & $89.3$ & $\textbf{87.1}$ & $82.5$ & $\textbf{90.2}$ & $28.1$M \\
\bottomrule
\end{tabular}
}
\end{center}
\caption{Results on the MPII validation set (PCKh@$0.5$). The input size is $256\times256$.} \vspace{-0.1in}
\label{tab:mpii}
\end{table}

\begin{table*}
\begin{center}
\resizebox{1.8\columnwidth}{!}{
\begin{tabular}{l|c|c|c|cccccc}
\toprule
Method & Input size & \#Params & GFLOPs & $AP$ & $AP^{50}$ & $AP^{75}$ & $AP^M$ & $AP^L$ & $AR$ \\ \midrule
G-RMI~\cite{rmi} & $353\times257$ & $42.6$M & $57$ & $64.9$ & $85.5$ & $71.3$ & $62.3$ & $70$ & $69.7$ \\
Integral Pose Regression~\cite{sun2018integral} & $256\times256$ & $45.0$M & $11$ & $67.8$ & $88.2$ & $74.8$ & $63.9$ & $74$ & - \\
CPN~\cite{cvpr18-cascaded} & $384\times288$ & - & - & $72.1$ & $91.4$ & $80$ & $68.7$ & $77.2$ & $78.5$ \\
RMPE~\cite{rmpe} & $320\times256$ & $28.1$M & $26.7$ & $72.3$ & $89.2$ & $79.1$ & $68$ & $78.6$ & - \\
SimpleBaseline-Res$152$\cite{ECCV18-xiao-simple} & $384\times288$ & $68.6$M & $35.6$ & $73.7$ & $91.9$ & $81.1$ & $70.3$ & $80$ & $79$ \\

HRNet-W$48$~\cite{20201207-human_pose-hrnet} & $256\times192$ & $63.6$M & $14.6$ & $74.2$ & $92.4$ & $82.4$ & $70.9$ & $79.7$ & $79.5$ \\

HRNet-W$32$~\cite{20201207-human_pose-hrnet} & $384\times288$ & $28.5$M & $16$ & $74.9$ & $92.5$ & $82.8$ & $71.3$ & $80.9$ & $80.1$ \\
TransPose-H-A$6$~\cite{yang2020transpose} & $256\times192$ & $17.5$M & $21.8$ & $75.0$ & $92.2$ & $82.3$ & $71.3$ & $81.1$ & $80.1$ \\
HRNet-W$48$~\cite{20201207-human_pose-hrnet} & $384\times288$ & $63.6$M & $32.9$ & $75.5$ & $\textbf{92.5}$ & $83.3$ & $71.9$ & $81.5$ & $80.5$ \\
\midrule
TokenPose-S-v$2$ & $256\times192$ & $6.2$M & $11.6$ & $73.1$ & $91.4$ & $80.7$ & $69.7$ & $79.0$ & $78.3$  \\
TokenPose-B & $256\times192$ & $13.5$M & $5.7$ & $74.0$ & $91.9$ & $81.5$ & $70.6$ & $79.8$ & $79.1$ \\
TokenPose-L/D$6$ & $256\times192$ & \multicolumn{1}{c|}{$20.8$M} & \multicolumn{1}{c|}{$9.1$} & $74.9$ & $92.1$ & $82.4$ & $71.5$ & $80.9$ & $80.0$ \\
TokenPose-L/D$24$ & $256\times192$ & \multicolumn{1}{c|}{$27.5$M} & \multicolumn{1}{c|}{$11.0$} & $75.1$  & $92.1$ & $82.5$ & $71.7$ & $81.1$ & $80.2$  \\

\textbf{TokenPose-L/D}$\textbf{24}$ & $384\times288$ & \multicolumn{1}{c|}{$29.8$M} & \multicolumn{1}{c|}{$22.1$} & $\textbf{75.9}$  & $92.3$ & $\textbf{83.4}$  & $\textbf{72.2}$ & $\textbf{82.1}$ &  $\textbf{80.8}$  \\
\bottomrule
\end{tabular}
}
\end{center}
\caption{Comparisons with state-of-the-art CNN-based models on the COCO test-dev set.}
\label{tab:test-dev}
\end{table*}

\subsection{MPII Human Pose Estimation}
\paragraph{Dataset \& Evaluation metric.}The MPII Human Pose dataset~\cite{mpii} contains images with full-body pose annotations obtained from various real-world activities. There are 40k person samples with 16 joints labels in the MPII dataset. In addition, the data augmentation is the same to that on the COCO dataset, except that the input images are cropped to $256\times256$. The head-normalized probability of correct keypoint (PCKh)~\cite{mpii} score is adopted for evaluation.

\paragraph{Results on the validation set.} We follow the testing procedure in HRNet~\cite{20201207-human_pose-hrnet}. The PCKh@$0.5$ results of some top-performed methods are presented in Table~\ref{tab:mpii}. All the experiments are conducted with the input image size $256\times256$. It's shown that our proposed TokenPose achieves competitive performance while being more lightweight.

\begin{table}[]
\begin{center}
\resizebox{\columnwidth}{!}{
\begin{tabular}{l|c|c|c}
\toprule
Model & Token fusion & AP & \#Params \\ \midrule
TokenPose-S & \ding{55} & $73.5$ & $6.2$M \\
TokenPose-S & \Checkmark  & $72.6$ & $6.7$M \\
TokenPose-L+/D$12$ & \ding{55} & $75.3$ & $35.8$M \\
TokenPose-L+/D$12$ & \Checkmark  & $75.5$ & $38.2$M \\ \bottomrule
\end{tabular}
}
\end{center}
\caption{The effects of keypoint token fusion for different models. The input image size is $256\times192$.}\vspace*{-0.01in}
\label{tab:fusion}
\end{table}

\subsection{Ablation Study}
\paragraph{Keypoint token fusion.} Intermediate supervision is widely-used to help model training and improve the heatmap estimation quality especially when networks become very deep~\cite{20201207-HumanPose-hourglass,wei2016convolutional,tompson2015efficientobject,belagiannis2017recurrent}. Similarly, we propose to concatenate keypoint tokens outputted by different layers of the Transformer encoder correspondingly, namely `keypoint token fusion', to help model training.

Taking TokenPose-L+/D$12$ with $12$ Transformer layers as an example, the keypoint tokens output in the $4$th, $8$th and $12$th layers are concatenated correspondingly. The resulting three times longer keypoint tokens are then sent into the MLP head to obtain the final heatmaps.   

We report the results of TokenPose-S and TokenPose-L+/D$12$ with and without keypoint token fusion in Table~\ref{tab:fusion}. For TokenPose-L+/D$12$, using keypoint token fusion improves the result by $0.2$ AP. However, for small variant like TokenPose-S, it causes performance degradation instead.

For TokenPose-Large with keypoint token fusion, we find the lower Transformer layers provide more meaningful evidence than the higher layers to understand the interaction process. We attribute this to the token fusion, which enables the final keypoint representation to directly exploit the information from the early layers. And such a phenomenon does not appear in the TokenPose-Small model without token fusion, in which the attention interactions progressively show clear and meaningful attention process. We will further describe it in Sec.~\ref{visualization}.

 Note that keypoint token fusion is only used in TokenPose-L given its very deep and complex structure. 
 
\begin{table}[t]
\begin{center}
\resizebox{\columnwidth}{!}{
\begin{tabular}{c|c|c|cc}
\toprule
\multirow{2}{*}{Position embedding} & \multirow{2}{*}{\#Params} & \multirow{2}{*}{GFLOPs} & \multirow{2}{*}{AP} & \multirow{2}{*}{AR} \\
 &  &  &  &  \\ \hline
\ding{55} & $6.62$M & $2.07$ & $67.0$ & $73.4$ \\
Learnable & $6.67$M & $2.23$ & $71.4$ & $77.1$ \\
$2$D sine & $6.67$M & $2.23$ & $72.5$ & $78.0$ \\ \bottomrule
\end{tabular}
}
\end{center}
\caption{Results for various positional encoding strategies for TokenPose-S-v$1$. The input image size is $256\times192$.}\vspace*{-0.01in}
\label{tab:pe}
\end{table}

\begin{figure*}
    \centering
    \vspace{-1em}
    \includegraphics[width=1.0\textwidth]{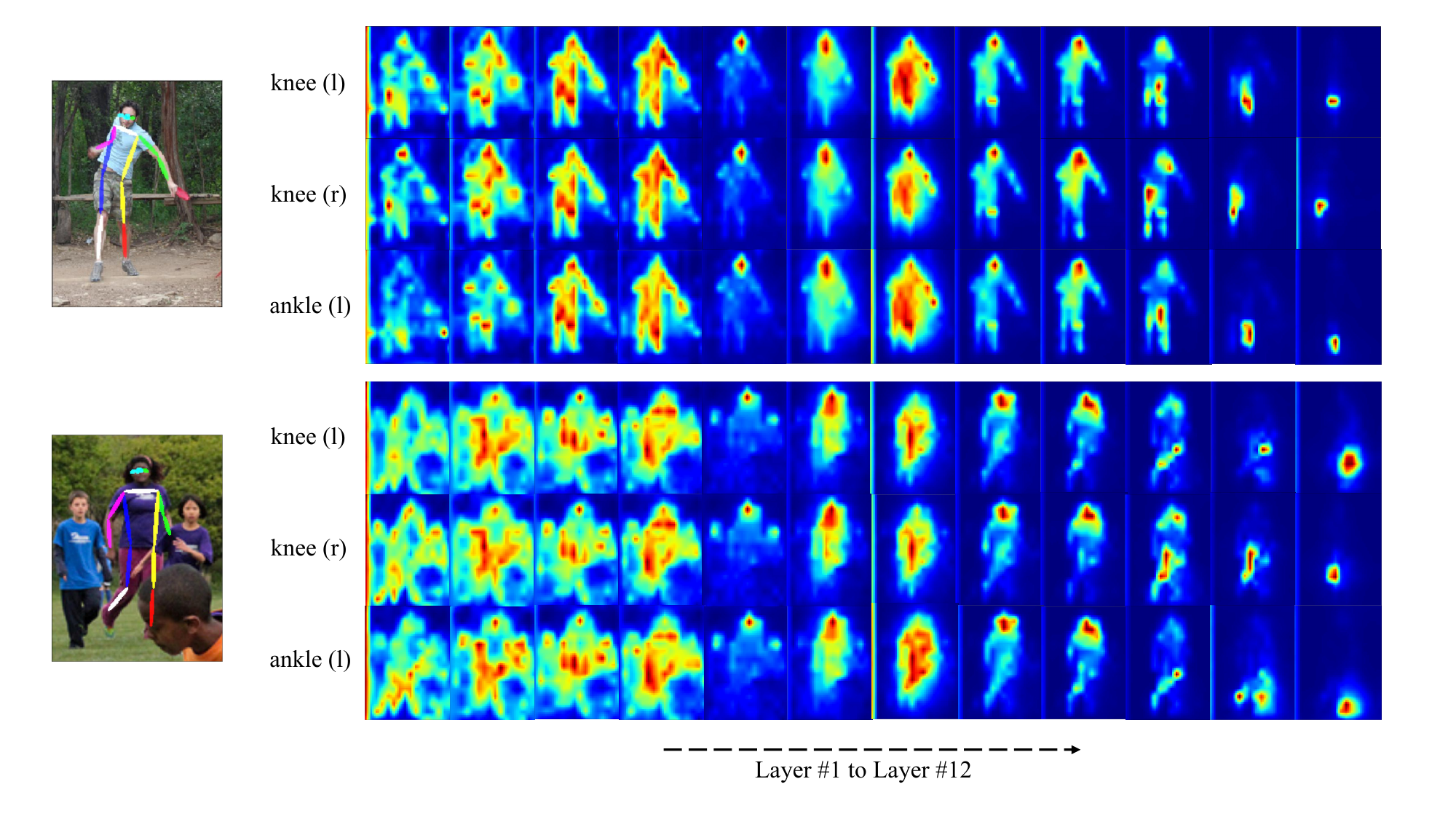}
    \vspace{-2.2em}
    \caption{Visualization of the attention maps between keypoint tokens (e.g., \emph{nose}, \emph{elbow(l)}, and \emph{elbow(r)}, etc.) and visual tokens in different layers of TokenPose-S, which consists of $12$ Transformer layers. Note that we transform all visual token into its corresponding patch areas in the image. Redder color areas mean that the given type of keypoint has higher attention at  these patches/visual tokens. The examples shown above and below are non-occluded and occluded cases, respectively.}
    \label{fig:appearance}
\end{figure*}

\begin{figure}[h]
    \centering
    \hspace{-2.5em}
    \includegraphics[width=0.52\textwidth]{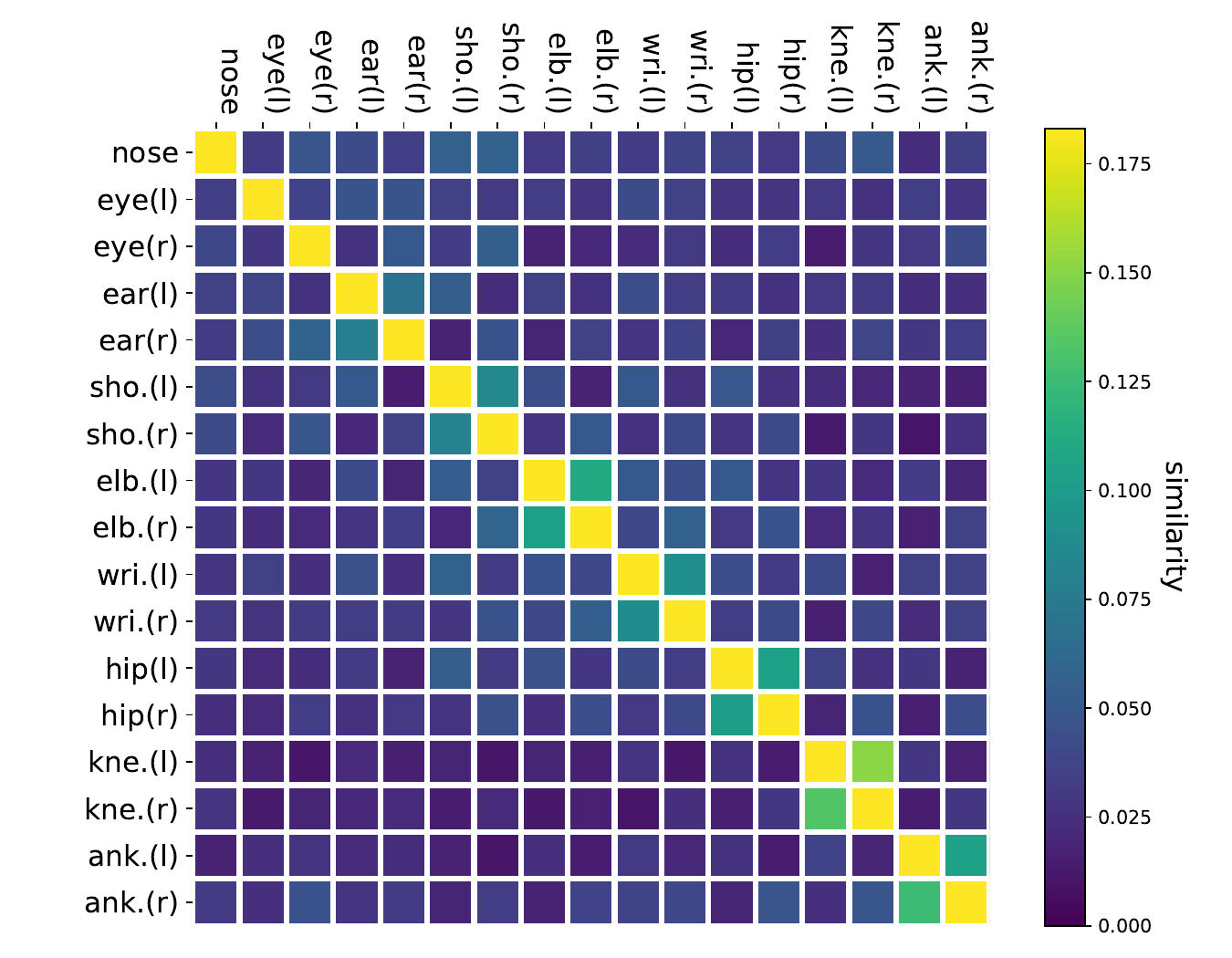}
    \hspace{-2.4em}
    \caption{The inner product matrix of the learned keypoint tokens. We take the keypoint tokens that are fed into the first Transformer layer, compute their inner product matrix, scale them by $\sqrt{d}$, and use softmax to normalize them at columns. Thus each row can represent the learned prior constraint relationships for a given type of keypoint with other ones.}\vspace{-0.8em}
    \label{fig:prior}
\end{figure}
 
\paragraph{Position embedding.} Keypoint localization is a position-sensitive vision task. To illustrate the effect of position embedding, we conduct experiments based on TokenPose-S-v$1$ with different position embedding types (i.e., no position embedding, $2$D sine and learnable position embedding). As Table~\ref{tab:pe} shown, employing position embedding significantly improves the performance by $5.5$ AP at most. In particular, $2$D sine position embedding performs better than learnable position embedding, which is as expected since the $2$D spatial information is required for predicting heatmaps.

\paragraph{Scaling.} Model scaling is a widely-used method to boost model performance, including width-wise ~\cite{transformer,devlin2018bert} scaling and depth-wise scaling~\cite{brown2020language,shoeybi2019training}. As shown in Table~\ref{tab:scaling}, both increasing depth and width help improve the results. 

\subsection{Visualization}
\label{visualization}

\begin{figure*}[h]
    \centering
    \vspace{-2em}
    \includegraphics[width=0.9\textwidth]{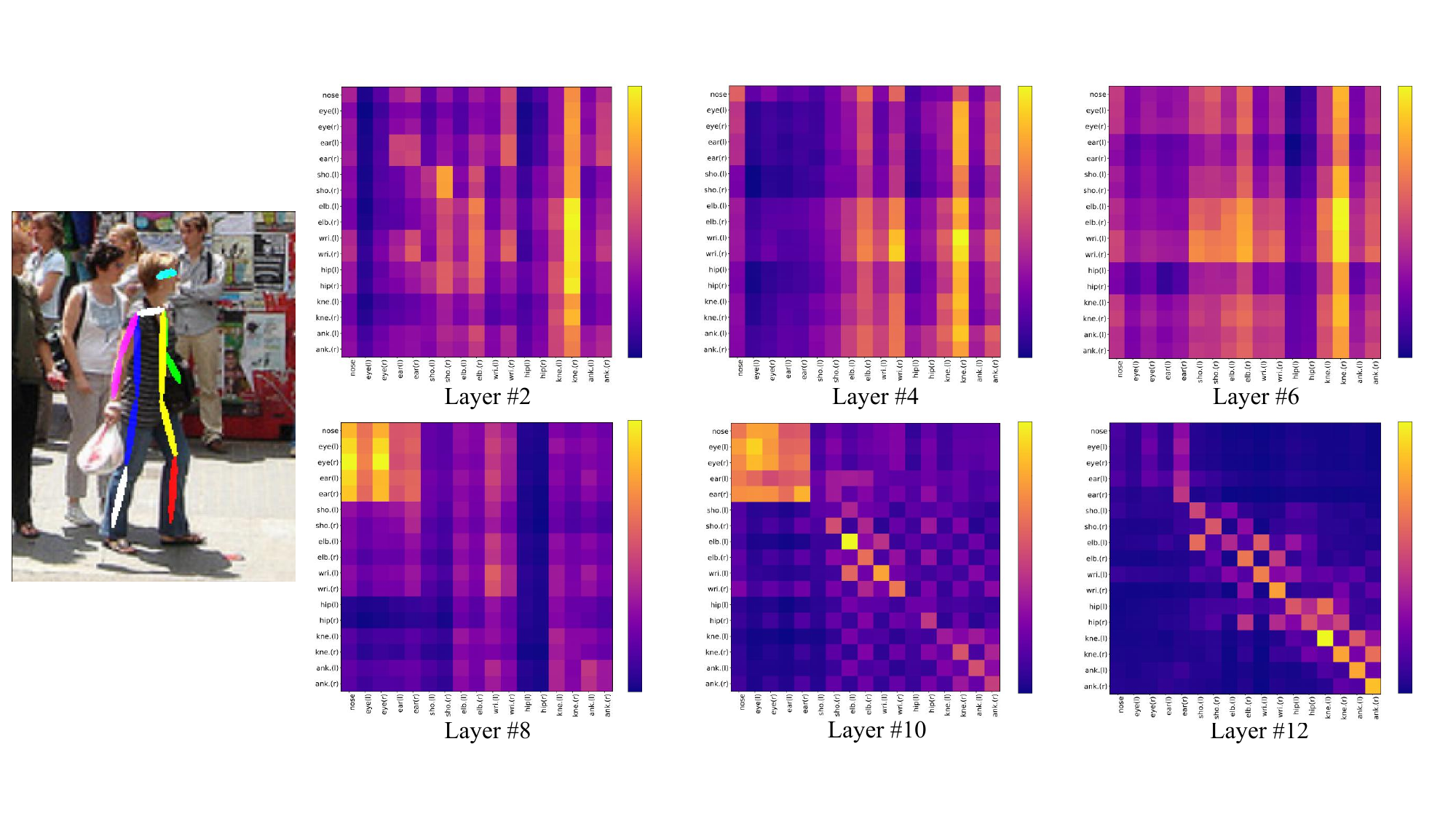}
    \vspace{-0.7cm}
    \caption{The attention interactions between keypoint tokens in the $2$nd, $4$th, $6$th, $8$th, $10$th and $12$th Transformer layers of TokenPose-S.}
    \label{fig:constraint}
    \vspace{-0.8em}
\end{figure*}

To illustrate how the proposed TokenPose explicitly utilizes visual cue and constraint cue between parts to localize keypoints, we visualize the details during inference. We observe that a single model has similar behaviors for most common examples. We randomly choose some samples from the COCO validation set and visualize the details in Figure~\ref{fig:appearance} and Figure~\ref{fig:constraint}. 

\paragraph{Appearance cue.}
 We visualize the attention maps between keypoint tokens and visual tokens of different Transformer layers in Figure~\ref{fig:appearance}. The attention maps are formed based on the attention scores between keypoint tokens and visual tokens. Note, we reshape the $1$D sequence of attention scores according to their original space positions for the visualization. 

We choose two images for comparisons in Figure~\ref{fig:appearance}. As we can see, with the layer depth increasing, what the keypoint tokens capture is gradually from the whole body appearance cues to more precise local part cues. In the first few layers, multiple crowded persons may simultaneously give appearance cues as interference, but the model can progressively attend to the target person.  
In the subsequent layers, different types of keypoint tokens attend to their adjacent keypoints and the joints with high confidence evidence. 

When inferring the occluded keypoints, the model behaves differently. As shown in Figure~\ref{fig:appearance}, we notice that the occluded \emph{left-ankle} keypoint token pays higher attention to its symmetric joint (i.e., \emph{right-ankle}) to obtain more clues.

\begin{table}[]
\begin{center}
\centering
\resizebox{1\columnwidth}{!}{
\begin{tabular}{l|cc}
\toprule
\multirow{2}{*}{Keypoint} & \multicolumn{2}{c}{Constraint} \\ \cline{2-3} 
 & Top-$1$ & Top-$2$ \\ \midrule
left shoulder & left elbow ($0.026$) & right shoulder ($0.012$) \\
left hip & right hip ($0.037$) & left knee ($0.037$) \\
right ankle & right knee ($0.023$) & left ankle ($0.014$) \\
nose & left eye ($0.016$) & right eye ($0.016$) \\
right wrist & right elbow ($0.012$) & left wrist ($0.011$) \\ \bottomrule
\end{tabular}
}
\end{center}
\caption{Top-$2$ constraints with regard to some keypoints for a randomly chosen sample. The values in parentheses represent the attention scores obtained from the final self-attention layer.}
\label{tab:keycue}
\end{table}

\paragraph{Keypoint constraints cue.} The attention maps of keypoint tokens in the $2$nd, $4$th, $6$th, $8$th, $10$th and $12$th self-attention layers are visualized in Figure~\ref{fig:constraint}. In the first few layers, each keypoint pays attention to almost all other ones to construct global context. As the network goes deeper, each keypoint tends to mostly rely on several parts to yield the final prediction.

Specifically, we show top-$2$ constraints of some typical keypoints based on the final self-attention layer in Table~\ref{tab:keycue}. In particular, we observe that the top-$2$ constraints tends to be the \emph{adjacent} and \emph{symmetric constraint} of the target keypoint, which also conforms to the human visual system. For instance, predicting the \emph{right wrist} mostly focuses on the constraints from the \emph{right elbow} and \emph{left wrist}, corresponding to its \emph{adjacent} and \emph{symmetric constraints} respectively.

\paragraph{Keypoint tokens learn prior knowledge from data.} In proposed TokenPose, the input \verb|[keypoint]| tokens which are taken as input to the first Transformer layer are totally learnable parameters. Such knowledge is related to the bias from the whole training dataset but independent of any specific image. During inference it will be exploited to facilitate the model to decode visual information from a concrete image and further make predictions. 

We point out that such \verb|[keypoint]| tokens act like object queries in DETR~\cite{detr}, in which each query slot finally has learned prior preference from data to specialize on certain areas and box sizes. In our settings, the input \verb|[keypoint]| tokens learn statistical relevance between keypoints from the dataset, serving as prior knowledge.

To show what information is encoded in these input keypoint tokens, we calculate the inner product matrix of them. After being scaled and normalized, the matrix is visualized in Figure~\ref{fig:prior}. We can see that one tends to be highly similar to its symmetric keypoints or adjacent keypoints. For instance, \emph{left hip} is mostly related to \emph{right hip} and \emph{left shoulder} with similarity score $0.104$ and $0.054$ respectively. Such finding conforms to our common sense and reveals what the model learns. We also notice there is a work~\cite{Tang_2019_CVPR} which analyzes the statistic distributions between joints by computing the mutual information from MPII dataset annotation. In contrast, our model is able to automatically learn prior knowledge from training data and explicitly encode it in the input \verb|[keypoint]| tokens. 

\section{Conclusion}
In this paper, we propose a novel token-based presentation for human pose estimation, namely TokenPose. In particular, we split the image into patches to yield visual tokens and represent keypoint entities into token embeddings. This way, the proposed TokenPose is able to explicitly capture appearance cues and constraint cues by the self-attention interaction. We show that a low-capacity pure Transformer architecture without any pre-training can also work well. Besides, the hybrid architectures achieve competitive results compared to the state-of-the-art CNN-based methods at a much lower computational cost.

\section*{Acknowledgments}
This paper is supported in part by the National Key R\&D Plan of the Ministry of Science and Technology (Project No. 2020AAA0104400), and in part by the National Key Research and Development Program of China under Grant 2018YFB1800204, the National Natural Science Foundation of China under Grant 61771273, the R\&D Program of Shenzhen under Grant JCYJ20180508152204044, and in part by the National Natural Science Foundation of China under Grant 61773117.

{\small
\bibliographystyle{ieee_fullname}
\bibliography{main}
}

\end{document}